\documentclass[11pt]{article}

\usepackage[final]{acl}

\usepackage{times}
\usepackage{latexsym}
\usepackage{amsmath}
\usepackage{linguex}

\usepackage[T1]{fontenc}

\usepackage[utf8]{inputenc}

\usepackage{microtype}

\usepackage{inconsolata}

\usepackage{graphicx}
\usepackage{enumitem}
\usepackage{booktabs}

\title{Collocational bootstrapping: A hypothesis about the learning of subject-verb agreement in humans and neural networks}

\author{
  Claire Hobbs \\
  Cognitive Science Program\\
  Yale University \\
  \texttt{claire.hobbs@yale.edu} \\\And
  R. Thomas McCoy \\
  Dept. of Linguistics \& Wu Tsai Institute\\
  Yale University\\
  \texttt{tom.mccoy@yale.edu}
  \\}

\begin{document}
\maketitle
\begin{abstract}
In what ways might statistical signals in linguistic input assist with the acquisition of syntax?
Here we hypothesize a mechanism called collocational bootstrapping, in which regularities in word co-occurrence patterns can provide cues to syntactic dependencies. We investigate whether this mechanism can support the acquisition of English subject-verb agreement.
First, we simulate language acquisition by training neural networks on synthetic datasets that vary in how predictable their subject-verb pairings are. We find that there is a range of variability levels at which these statistical learners robustly learn subject-verb agreement.
We then analyze the variability of subject-verb pairings in child-directed language, and we find that the variability in such data falls within the range that supported robust generalization in our computational simulations. 
Taken together, these results suggest that collocational bootstrapping is a viable learning strategy for the type of input that children receive. 
\end{abstract}

\section{Introduction}

The sentences that we encounter do not come annotated with explicit syntax trees. How, then, do children acquire a language's syntax?
Some proposals postulate innate predispositions that might guide children toward particular structural analyses \cite[e.g.,][]{chomsky_1965}.
Other proposals point to ways in which non-syntactic aspects of the data, such as semantics \cite{wexler1980formal} or prosody \cite{morgan1996signal}, might provide helpful cues 
from which the learner can ``bootstrap'' syntactic structure. For instance, prosodic boundaries tend to coincide with syntactic boundaries such that observable prosody might point toward accurate analyses of unobservable syntax.

Recent advances in artificial intelligence have given prominence to one particular type of data-driven cue: statistical properties of linguistic strings. Neural network models trained to capture the statistical properties of corpora perform well on tests that target syntactic phenomena such as filler-gap dependencies \cite{wilcox2024using}, negative polarity items \cite{jumelethupkes2018language}, and subject-auxiliary inversion \cite{mueller2022coloring}. These systems do not have explicit syntactic predispositions built into them, so their strong syntactic abilities suggest that naturalistic text possesses statistical cues from which much of syntax can be inferred. 
This evidence \textit{that} statistical properties can pave the way to syntax raises the question of \textit{how} they might do so.

In this work, we propose \textbf{collocational bootstrapping} as a hypothesis for one specific way in which statistical cues could contribute to syntactic acquisition. Under this hypothesis, syntactic structure can be inferred from trends regarding which words frequently co-occur. As a case study, we consider English subject-verb agreement, the phenomenon in which a verb must have the same grammatical number as its subject (e.g., \textit{the dogs bark} is grammatical, but \textit{the dogs barks} is not). One challenge for acquiring subject-verb agreement is that there are (at least) two potential rules that could explain most examples of this phenomenon:

\ex. \textsc{Agree-Subject}: A verb should agree with its subject.

\ex. \textsc{Agree-Recent}: A verb should agree with the closest preceding noun.

We can tell that \textsc{Agree-Subject} is the correct rule by considering sentences where these rules make different predictions;
e.g., when choosing a verb for the sentence \textit{the dogs in the park [bark/barks]}, \textsc{Agree-Subject} would correctly choose \textit{bark} while \textsc{Agree-Recent} would incorrectly choose \textit{barks}. 
However, for most naturally-occurring sentences, a verb's subject is also the most recent noun, meaning that it may be challenging for learners to identify which rule is correct.

The proposed mechanism of collocational bootstrapping (described in more detail in Section~\ref{sec:collocational_bootstrapping}) provides one way to select between these two rules even in the absence of direct disambiguating examples such as \textit{the dogs in the park bark}. Under the collocational bootstrapping hypothesis, learners leverage information about word co-occurrence as a window into syntactic dependencies. E.g., given \textit{the dog on the couch barks}, a learner could infer that there is more likely to be a dependency between \textit{dog} and \textit{barks} than between \textit{couch} and \textit{barks} because \textit{dog} is more likely to co-occur with \textit{barks} than \textit{couch} is. 
After inferring many such potential dependencies, the learner then abstracts away from the specific words that are involved to recognize the abstract syntactic configurations that are truly at the heart of the dependency.

To investigate whether collocational bootstrapping is a viable strategy, we trained multiple neural network language models on synthetically-generated datasets. Because collocational bootstrapping depends on associations between subjects and verbs, we varied the extent to which a subject could be predicted from its verb. Specifically, subjects were sampled from Zipfian distributions \cite{zipf_1949} where the probability of a verb's $r^{\text{th}}$ most frequent subject is proportional to $1/r^\alpha$; varying the parameter $\alpha$ modulates how predictable the subject is given the verb.
Critically, the models' training sets were constrained to be fully ambiguous between \textsc{Agree-Subject} and \textsc{Agree-Recent} (e.g., using sentences like \textit{the dog in the park barks}), but the systems were then evaluated on sentences that disambiguated these rules. 


We find that subject-verb co-occurrence statistics have a substantial effect on how well the models learn subject-verb agreement; there are some statistical settings (namely, when $\alpha \approx 1.4$, yielding moderate variability) where the models successfully learn \textsc{Agree-Subject} and others where they do not (namely, when $\alpha$ is very low---producing highly variable data---or very high---producing highly predictable data). 
The fact that certain statistical configurations support effective generalization supports the collocational bootstrapping hypothesis. 
Given that collocational bootstrapping is only effective in certain statistical settings, we next perform a corpus analysis of a dataset of child-directed language to see whether children’s input has the properties that supported success in our simulations. We find preliminary evidence that child-directed language indeed has the requisite properties.

Overall, our neural-network experiments provide a proof of concept showing that collocational bootstrapping can guide a learner to accurate syntactic analyses, and our corpus analysis suggests that children’s input has the statistical properties that make collocational bootstrapping effective. This work is a step toward understanding how quantitative aspects of a learner’s input can support the learning of abstract, qualitative syntactic phenomena.\footnote{Our code is available on GitHub: \url{https://github.com/ClaireHobbs/collocational-bootstrapping}.}


\section{Background and Related Work}\label{sec:background}

\paragraph{Bootstrapping in language acquisition:} Several mechanisms have been proposed by which learners might infer aspects of syntax from non-syntactic information such as prosody \cite{morgan1996signal} or meaning 
\cite{wexler1980formal,pinker1984language,abend2017bootstrapping,yedetore2024semantic}.
The most relevant prior proposal is distributional bootstrapping, in which syntactic categories can be inferred from distributional properties---e.g., words occurring in similar contexts likely belong to the same part of speech \cite{maratsos1980internal,finch1992bootstrapping,mintz2003frequent}. Like distributional bootstrapping, collocational bootstrapping leverages distributional properties of words, but it is a strategy for acquiring relationships between words rather than word categories. 
Another proposal that is potentially related to collocational bootstrapping is semantic bootstrapping \cite{wexler1980formal}; see Section~\ref{sec:discussion} for discussion. 
The various types of bootstrapping are not mutually exclusive---children might use many or all of them.

\paragraph{Subject-verb agreement in neural networks:} A substantial body of work has investigated whether neural networks can learn English subject-verb agreement \cite{elman1991distributed,linzen_2016}. Such networks have been found to be capable of robustly learning subject-verb agreement from naturalistic text \cite{kuncoro2018lstms,gulordava_2018,goldberg_2019,wei_2021}.
In this work, we train networks on controlled synthetic data to analyze what statistical properties of corpora might be supporting such learning. Our approach shares with \citet{wei_2021} the strategy of training neural networks on corpora that vary in controlled ways, but we investigate a different factor (namely the predictability of the subject given the verb, as opposed to word frequency).

\paragraph{Distributional cues to syntax:} Both the acquisition literature and the computational literature have discussed what properties of a learner's input might support the acquisition of syntax. Properties discussed include the presence of sentences that might directly disambiguate competing hypotheses \cite{pullum2002empirical,mulligan2021structure}, the presence of one phenomenon that might be helpful for acquiring different phenomena \cite{pearl2016role,patil2024filtered,misra2024language, yang2026unifiedassessmentpovertystimulus}, the semantic features of a word's arguments \cite{misra2024generating}, the statistical properties of function words \cite{yang2026functionwordsstatisticalcues}, the frequencies of particular words \cite{wei_2021,leong2024testing}, the diversity and complexity of observed syntactic structures \cite{qin2025data}, 
and the frequencies of syntactic configurations \cite{wonnacott2008acquiring,yang2016price}. 
We instead study the distributional feature of variability in word co-occurrence statistics.

\paragraph{The role of variability in learning:} Across domains of cognition, the tradeoff between predictability and variability is a central tension for learning \cite{raviv_2022}: predictable input can support faster learning, while variable input supports more abstract generalizations.
Our work applies this idea to the learning of English subject-verb agreement by investigating variability in subject-verb pairings. 
The most relevant prior papers are \citet{gomez2002variability} and \citet{onnis2004variability}.
In both, human participants were shown strings of the form \textit{aXb}, where the first and third elements had an agreement dependency (akin to subject-verb agreement), and the \textit{X} elements were arbitrary. These papers found that people can learn such patterns more readily when the set of possible \textit{X} elements is larger, showing that variability in intervening elements can support learning of nonadjacent syntactic dependencies.

Our work differs from these papers in that we investigate variability in subject-verb pairings, rather than variability in the intervening material. Further, we test artificial neural networks rather than humans. Finally, while these prior papers achieved greater variability by increasing the number of possible \textit{X} entities, we held the relevant sets constant and varied only the frequencies of their elements. This choice means that our conditions differ only quantitatively---there are no qualitative differences regarding which pairings are present (except in one case, the $\alpha\to\infty$ case). 



\section{The Collocational Bootstrapping Hypothesis}\label{sec:collocational_bootstrapping}

Words are not distributed uniformly in natural language use. Most relevantly for this paper, a given verb is much more likely to have some subjects than others due to the meanings that people are likely to express (e.g., \textit{the dog barked} is much more likely than \textit{the potato barked}).
We hypothesize that co-occurrence properties are likely to be especially systematic for words that share a syntactic dependency, such that learners could leverage co-occurrence information to help learn aspects of syntax, such as subject-verb agreement.

As discussed above, there is a tension between predictability and variability. If a given verb always had the same subject, it would be easy for the learner to recognize which noun the verb is paired with such that \textsc{Agree-Subject} can be selected over \textsc{Agree-Recent}. However, the learner in this setting might simply memorize the few subject-verb pairings it has seen, thereby failing to generalize to novel pairings. At the other extreme, if subjects are sampled uniformly, this high degree of variability should support generalization to novel subject-verb pairings, but it would not provide any systematic co-occurrence information that would help point to which noun is the verb's subject. 

Given this tension, the key question underlying our first experiment is whether there exists a level of variability that supports correct generalization of subject-verb agreement---a level that is predictable enough to make subject-verb associations apparent yet variable enough to support generalization to novel subject-verb pairs.
To get at this question, we use simulations with neural networks trained on simple, synthetic grammars. Using synthetic grammars enables us to fully control and understand which cues are available to the learners so that we can isolate the statistical factors we have highlighted, in the same spirit as other connectionist work that similarly analyzes how neural networks generalize in simple, controlled settings \cite{elman1990finding,elman1991distributed,frank2013acquisition,mccoy2020does}.


\begin{table*}[t]
\centering
\small
\setlength{\tabcolsep}{8pt}  
\begin{tabular}{lll}
\toprule
\textbf{Template} & \textbf{Example (Singular)} & \textbf{Example (Plural)} \\
\midrule
Det N V & the singer sings & the singers sing \\
Det N PP V & the singer by the dancer sings & the singers by the dancers sing \\
PP Det N V & by the dancer the singer sings & by the dancers the singers sing \\
PP Det N PP V & by the swimmer the singer by the dancer sings & by the swimmers the singers by the dancers sing \\
\bottomrule
\end{tabular}
\caption{Templates used in training set generation. Det = determiner, N = noun, PP = prepositional phrase, V = verb.}
\label{tab:sentence_templates}
\end{table*}

\section{Experiment 1: Neural Networks}

Neural networks allow us to simulate language acquisition in a statistical learner that lacks explicit predispositions for specific syntactic structures. To study the effect of certain statistical properties on learnability, we can train a neural language model on synthetic datasets in which we vary these properties in controlled ways. In our case, to modulate the level of variability in subject-verb pairings, we sample pairings from Zipfian distributions (defined by the equation below) that vary the parameter $\alpha$; note that $\alpha$ is a free parameter while $K$ is a normalizing constant whose value is fully determined by the need for the set of $f(r)$ values to sum to 1:

\begin{equation}\label{eq:zipf}
f(r) = \frac{K}{r^{\alpha}}
\end{equation}
Our $\alpha$ values ranged from 0 to 3, with lower $\alpha$ values producing highly variable pairings and higher $\alpha$ values producing predictable pairings, and we included an $\alpha \to \infty$ scenario in which each subject was seen paired with only one verb. 
After training, we evaluated the model’s ability to generalize subject-verb agreement beyond its training data. 

This highly simplified setup creates a proof-of-concept test to determine whether there exist situations in which collocational bootstrapping would be an effective strategy for a statistical learner. Specifically, there is no guarantee that collocational bootstrapping can ever succeed because it may be that all training sets are either too predictable to support abstract generalization or too variable to provide a clear statistical signal (see Section~\ref{sec:collocational_bootstrapping}). By modulating $\alpha$, we investigate multiple levels of variability to see if there exist levels that resolve this tension between predictability and variability, such that conditions exist under which collocational bootstrapping can succeed.

\begin{table*}[t]
\centering
\small
\begin{tabular}{p{4cm} p{10cm}}
\toprule
\textbf{Condition} & \textbf{Example Minimal Pair} \\
\midrule
\addlinespace
SEEN, MATCH & Near the fisher the bridger near the miner [listens / listen] \\
UNSEEN, MATCH & Near the \underline{crafter} the \underline{twirler} near the \underline{singer} [lassos / lasso] \\
SEEN, MISMATCH & By the protectors the trader by the teachers [trades / trade] \\
UNSEEN, MISMATCH & By the \underline{climbers} the \underline{driver} by the \underline{writers} [hunts / hunt] \\
\addlinespace
\bottomrule
\end{tabular}
\caption{Example minimal pairs for a moderately variable condition ($\alpha$ = 1.5). Underlining indicates that the noun has not been seen in training data paired with this verb.}
\label{tab:example_pairs}
\end{table*}

\subsection{Data}

We created synthetic datasets containing 12,000 unique grammatically-correct sentences for every $\alpha$ value tested. Each sentence contained a subject (made of a determiner and a noun) and an intransitive verb, and it could optionally include a prepositional phrase before and/or after the subject.
This setup produced four sentence templates (Table~\ref{tab:sentence_templates}).
Within each sentence, all nouns had the same number: Either all were singular, or all were plural. This meant that the training set was always ambiguous between \textsc{Agree-Subject} and \textsc{Agree-Recent} as well as many other potential rules (e.g., \textsc{Agree-First}, in which a verb agrees with the first noun in the sentence). 
We enforced this ambiguity to isolate the hypothesized effect of collocational bootstrapping: Are co-occurrence patterns sufficient to disambiguate candidate agreement rules even in the absence of sentences that would directly disambiguate these rules? See Section~\ref{sec:discussion} for discussion of future directions that relax this strict ambiguity.

Our vocabulary included 40 nouns and 40 present-tense verbs (each of which could be singular or plural, creating a total of 80 nouns and 80 verbs). For reasons described at the end of this section, each noun in the vocabulary is derived from one of the verbs, creating noun/verb pairs such as \textit{orator}/\textit{orates}. We chose \textit{the} as the only determiner,
and we used \textit{by} and \textit{near} as our prepositions. 

To explain how sentences were sampled, we first give each noun stem and verb stem a numerical index from 0 to 39. To generate a sentence, first the verb was sampled uniformly from among the 80 options; call its index $i$.
A subject was then sampled from a truncated Zipfian distribution with parameter $\alpha$ that assigns the following unnormalized probability to each of the 40 nouns that can agree with verb $i$, denoting a given noun's index as $j$: $1/((j-i \mod 40) + 1)^\alpha$ if $j-i \mod 40 < 30$, else $0$. That is, for each verb, there are 10 nouns that are withheld from appearing as the verb's subject so that we can evaluate how the models generalize to novel subject-verb pairs.
If the sentence contained prepositional phrases, the prepositional object nouns were sampled uniformly from among the nouns with indices $i$ to $i + 29 \mod 40$ that have the same number as the subject. One effect of this setup is that the same 10 nouns that were withheld from appearing as the verb's subject were also withheld from appearing as prepositional objects for purposes of evaluation. 

\begin{figure}[t]
    \centering
    \includegraphics[width=\columnwidth]{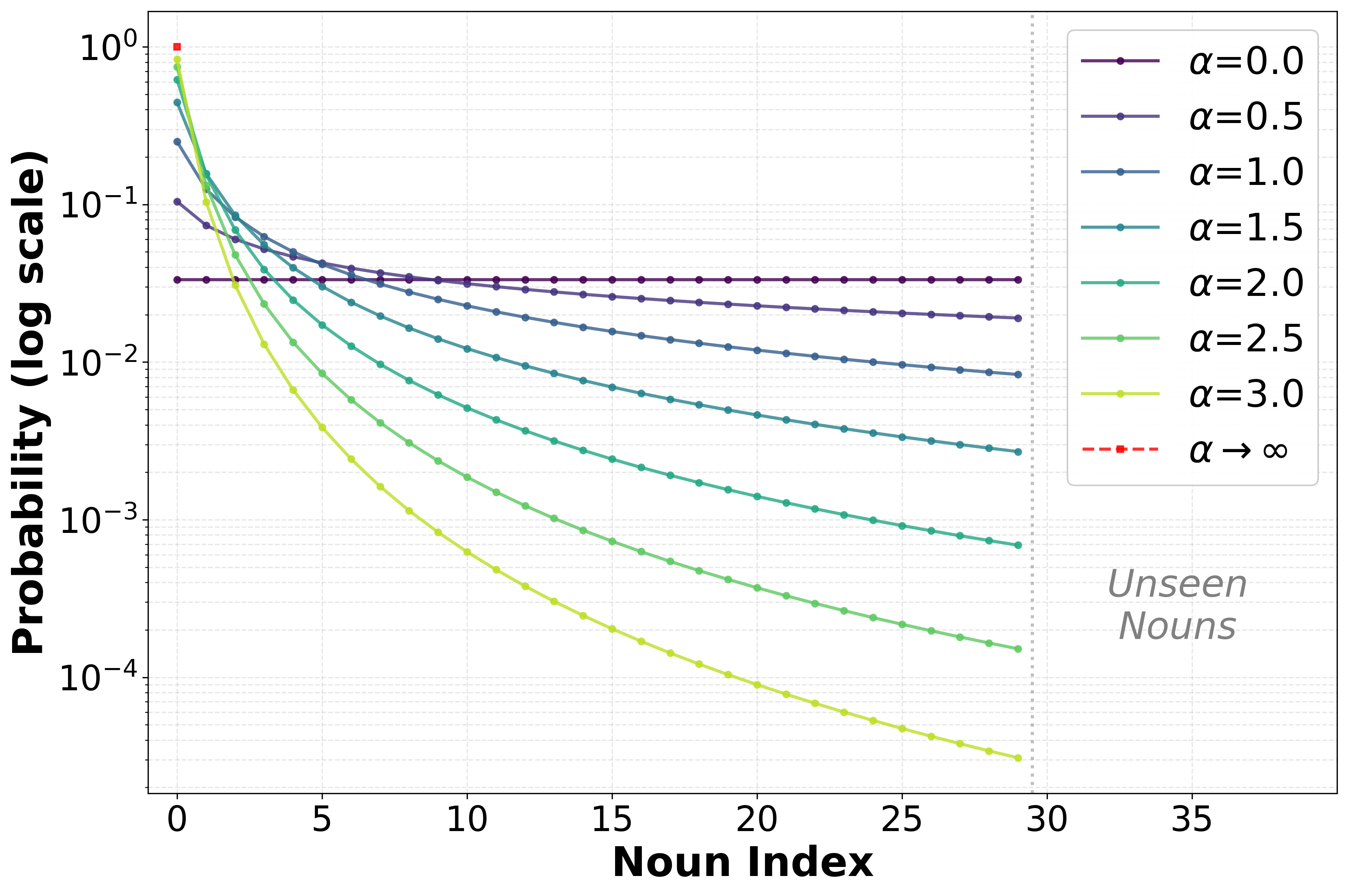}
    \caption{Noun probability distributions across $\alpha$ values (log scale). Lower $\alpha$ values produce flatter distributions with more uniform noun usage, while higher $\alpha$ values concentrate probability on fewer nouns. The distributions were truncated at the dotted line to leave some nouns unseen as the subjects of particular verbs.}
    \label{fig:noun_dist}
\end{figure}

By controlling $\alpha$, we can adjust the level of variability in the training data with respect to which subjects appeared with which verbs. When $\alpha$ = 0, the subjects were distributed uniformly, creating a dataset with the highest level of variability. At the other extreme, when $\alpha \to \infty$, each verb (e.g., \textit{orate}) had only one noun that ever appeared as its subject---specifically, the noun that was morphologically related to it (e.g., \textit{orator}). For in-between values of $\alpha$, the nouns available for verb pairings were spread across a truncated Zipfian distribution as defined above, with the most likely subject being the one that is morphologically related to the verb, and other nouns having probabilities that decrease following Equation~\ref{eq:zipf}. 
We used Zipfian distributions because many linguistic units have been empirically shown to follow them, including in child-directed language \cite{lavi-rotbain_2023}; see Section~\ref{sec:expt2_results} for evidence that subject-verb pairings follow Zipfian distributions in child-directed language. 
Figure~\ref{fig:noun_dist} shows the distribution of nouns at each selected $\alpha$ value, and sample training sentences can be found in Appendix~\ref{appendix:training_sentences}.


Note that our models do not have access to the spellings of words; they are presented with words as atomic tokens. Thus, they cannot leverage the morphological cues of noun inflection, verb inflection, and the relatedness of certain nouns and verbs (e.g., \textit{painter} and \textit{paint})---these morphological properties are included in the dataset to make the sentences easier for humans to reason about, but these properties play no role in the models' learning. 
Additionally, our dataset makes many simplifying assumptions compared to natural language use; see Section~\ref{sec:discussion} for discussion.

\subsection{Models}

We trained and evaluated 2-layer decoder-only Transformer language models \cite{vaswani_2017} in the style of GPT-2 \cite{radford_2019}, adapted from the nanoGPT implementation \cite{karpathy_2023}, which enables lightweight, research-oriented versions of GPT-2 to be trained from scratch. Our models used two transformer layers, each with four attention heads, an embedding size of 256, and approximately 1.6 million parameters. All code was developed using PyTorch.

\subsection{Training}

For each $\alpha$ value from 0.0 to 3.0 inclusive in increments of 0.1, as well as the case where $\alpha\to \infty$, we did 10 training runs with different random weight initializations. For each run, we generated a new set of 12,000 unique sentences and used a split of  80\% train / 10\% validation / 10\% test. We used AdamW \cite{kingma2015adam,loshchilov2018decoupled} with a learning rate of 0.0006 which remained fixed throughout training. 
The batch size was 32, and each training run used 300 batches per epoch for 4 epochs (1,200 total iterations). The validation loss was computed every 300 steps, and the model version with the best validation loss was saved. 
Training and validation losses tracked each other closely (see Figure~\ref{fig:loss} in the Appendix), indicating that overfitting was not a concern.

\begin{figure*}[t]
    \centering
    \includegraphics[width=0.7\linewidth]{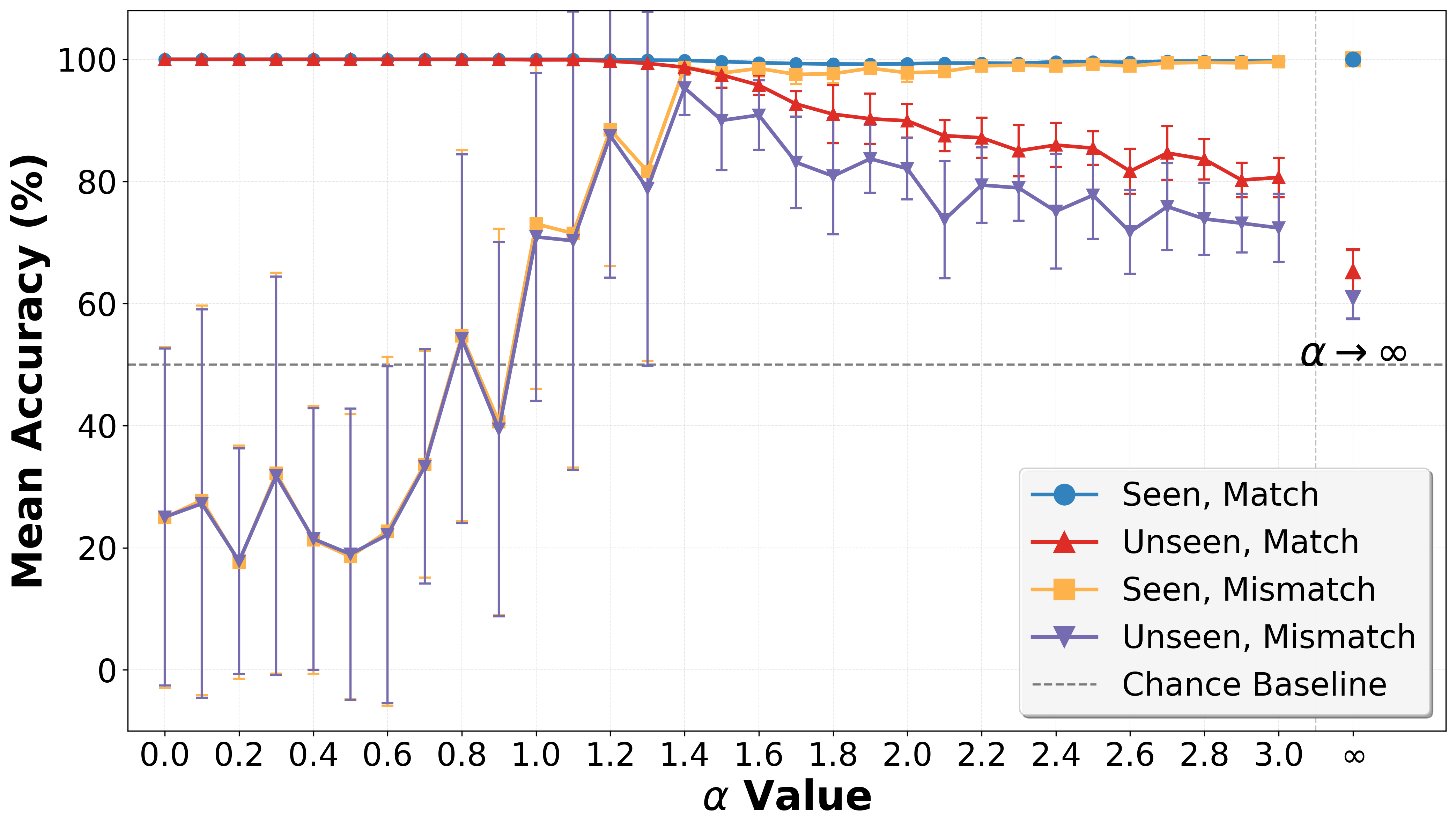}
    \caption{Model accuracy vs.\ Zipfian parameter $\alpha$ across four evaluation conditions. There is an optimal point where $\alpha=1.4$ at which models perform robustly in all test conditions. Error bars show one standard deviation.}
    \label{fig:accuracy}
\end{figure*}

\subsection{Evaluation and Results}
To evaluate model performance, we generated four sets of 1,000 minimal pair sentences \cite{marvin2018targeted}, each targeting a different testing condition. In each pair, the first sentence was grammatical, and the second was ungrammatical due to the verb not matching the subject's number. Sample minimal pairs are in Table~\ref{tab:example_pairs}.

We assessed each model's preferences by calculating the log probability it assigned to each sentence in a pair. For each pair, we considered the model to be correct if it assigned a higher log probability to the grammatical sentence than to the ungrammatical one, and we then computed the overall accuracy across the 1,000 pairs in each set.

The four sets varied in difficulty according to whether the subject-verb pairings and prepositional objects had appeared in the model’s training data (SEEN) or not (UNSEEN), and whether the grammatical number of prepositional object nouns matched that of the subject (MATCH) or not (MISMATCH). The number mismatches served as attractors to assess whether the model had learned \textsc{Agree-Subject} (which would identify the correct noun for the verb to agree with) or an incorrect strategy such as \textsc{Agree-Recent} or \textsc{Agree-First} (both of which would select the incorrect verb inflection). All minimal pairs shared a uniform syntactic structure, [PP Det N PP V], presenting the model with three competing nouns as possible agreement targets for the verb. Below, we define these four conditions in detail and present results for each. 

\paragraph{\textbf{SEEN, MATCH:}}
The sentences in this condition used subject-verb pairings the model encountered during training, with prepositional objects matching the subject's number. This condition presented the lowest difficulty for the models, serving primarily to verify that the models had successfully learned the patterns present in the training data. Across all $\alpha$ values, the models achieved 100\% or near 100\% accuracy as shown by the blue line in Figure~\ref{fig:accuracy}.

\paragraph{\textbf{UNSEEN, MATCH:}}
Here, we introduce a source of lexical difficulty. For a given verb, the subject and prepositional objects in the test sentences were  ones that had never appeared in the same sentence as that verb during training.
As in the previous condition, the prepositional objects matched the subject's number. Success in this condition requires the model to generalize across words of the same number---that is, to use distributional commonalities to form the classes of \textit{singular nouns} and \textit{plural nouns}, and to recognize that the same verb form applies to any member of that class. This type of generalization should be easiest when $\alpha$ is low, meaning that all the singular nouns have similar distributions and are therefore easier for the model to group together into a cohesive class, and similarly for the plurals. As expected, accuracy was high for low $\alpha$ values but began to decline when $\alpha \approx 1.4$ (see the red line in Figure~\ref{fig:accuracy}).
At high $\alpha$ values, the models appear to learn strong associations between specific subjects and verbs, preventing them from generalizing well to unseen ones.

\paragraph{\textbf{SEEN, MISMATCH:}}
This condition presents a different type of difficulty: conflicting cues about number agreement. If the model has incorrectly learned that the verb should agree either with the closest noun or with the first noun in the sentence---both of which would succeed for all sentences in the training set---it will now fail when tested with sentences in which the prepositional objects have a different grammatical number than the subject. High $\alpha$ values in the training data create greater predictability in subject-verb pairings, which we hypothesize would help models select \textsc{Agree-Subject} over other competing rules by making it easier to recognize which syntactic positions host the noun that the verb shares a syntactic dependency with. 
As shown by the yellow line in Figure~\ref{fig:accuracy}, results confirmed that the models perform poorly at low $\alpha$ values but with high accuracy
(approaching 100\%) as $\alpha$ increases.

\paragraph{\textbf{UNSEEN, MISMATCH:}}
Our final condition combines both sources of difficulty: novel subject-verb pairings and prepositional objects that have a different number from the subject (note that the prepositional object-verb pairings are also novel, as in the UNSEEN, MATCH condition). As above, we predict that low $\alpha$ values will prevent the model from generalizing because it will struggle to identify the correct agreement target among mismatching competitors, and that high $\alpha$ values will also cause the model to perform poorly, as it will struggle to generalize to unseen noun/verb pairings. What happens between the low and high $\alpha$ values is harder to predict. The critical question is whether there exists a ``sweet spot'' between extremes, where the model can handle both the UNSEEN and MISMATCH aspects of this condition. We find that there is indeed such a sweet spot (Figure~\ref{fig:accuracy}, purple line):  
The model showed poor performance at low and high $\alpha$ values, but there is a peak with near-perfect accuracy at intermediate values.


\subsection{Discussion}
The results show there is an ideal level of variability in subject-verb pairings in the training data that helps the model generalize robustly. 
Too much variation hinders the model from inferring the correct syntactic structure.
Too much predictability prevents the model from forming an abstract rule, such that it generalizes poorly to novel subject-verb pairings.
Between these extremes, when $\alpha \approx 1.4$, there is an optimal level of variability that supports robust generalization. This pattern demonstrates two key points. First, the fact that model performance varies with the level of variability indicates that these neural networks indeed use co-occurrence statistics to inform the learning of subject-verb agreement in the ways expected under the collocational bootstrapping hypothesis. 
Second, the existence of an optimal level at which we get robust generalization shows that, under the right conditions, collocational bootstrapping can be a viable learning strategy.

\section{Experiment 2: Analysis of CHILDES}\label{sec:childes}
In the previous experiment, we observed that when $\alpha \approx 1.4$, the synthetic training data contained a level of variability  that optimizes generalization. 
We now investigate whether a similar statistical signal is present in real-world data such that children could potentially leverage this signal to assist in learning subject-verb agreement. Toward this end, we consider the frequency of subject-verb pairings in a corpus of child-directed language.

\subsection{Data}

CHILDES, or the \textbf{Chi}ld \textbf{L}anguage \textbf{D}ata \textbf{E}xchange \textbf{S}ystem, is an open repository of transcripts and other supporting media containing conversations between children and their caretakers, which have been compiled and donated by researchers \cite{macwhinney_2000}. 
For this experiment, we used data from CHILDES participants tagged as English-language speakers. Because our goal is to understand the linguistic input that a child might receive (and not the utterances that the child produces),  we filtered the speakers to include only those speaking to children but not the children themselves.

We extracted all utterances spoken by an adult to children with ages from 0 to 96 months, resulting in a set of 4,739,189 utterances; see Appendix~\ref{appendix:data_cleaning} for more details about data filtering and cleaning.

We parsed the utterances using the spaCy dependency parser
\cite{honnibal_2020}. We extracted all pairings of a subject noun and the corresponding verb (the pairings characterized by the dependency type \textit{nsubj}), using lemmas for both subjects and verbs. 
This produced 2,802,071 subject-verb pairs.

\subsection{Zipfian Analysis}

We analyzed the subjects of the 100 most frequent verbs. We restricted ourselves to frequent verbs so that there would be sufficient data to achieve quantitatively meaningful results; all verbs in this set appeared at least 2,396 times. For each verb, we created a list of the subjects that co-occur with that verb, ranked by the number of times that the subject-verb pairing appeared. Next, we converted these subject counts to proportions and calculated the average proportion of the subjects at each rank across all verbs. That is, for each rank $r$ we computed $f_{\text{empirical}}(r)$---the average frequency of a verb's $r^{\text{th}}$ most common subject---as follows, where $\text{verb}_k$ is the $k$th most common verb, and $\text{subj}_{r,k}$ is the noun that occurs as the $r$th most common subject for $\text{verb}_k$:
\begin{equation}
f_{\text{empirical}}(r) = \frac{1}{100} \sum_{k=1}^{100} \frac{\text{count}(\text{subj}_{r,k}, \text{verb}_k)}{\text{count}(\text{verb}_k)}
\label{eq:empirical}
\end{equation}
This formula gives the empirical frequencies of verb-subject pairings, which we then sought to fit to the theoretical predictions of Zipf's Law:
\begin{equation}
f_{\text{theoretical}}(r,\alpha) = \frac{K}{r^\alpha}
\label{eq:theoretical}
\end{equation}

Zipf's Law has one free parameter $\alpha$ (note that $K$ is a normalizing constant, so it is fully determined by $\alpha$), so fitting $f_{\text{theoretical}}$ to $f_{\text{empirical}}$ amounted to finding the value of $\alpha$ that best fit the observed data. To do so, we tried all values of $\alpha$ ranging from 0 to 3.0 in increments of 0.01. For each $\alpha$ value, we computed the mean squared error (MSE) between $f_{\text{empirical}}$ and $f_{\text{theoretical}}$, defined as: 
\begin{equation}
\text{MSE}(\alpha) = \frac{1}{R} \sum_{r=1}^{R} \left(f_{\text{empirical}}(r) - f_{\text{theoretical}}(r,\alpha)\right)^2
\end{equation}
where $R$ is the number of ranks over which we computed the error. We then selected the $\alpha$ value that minimized $\text{MSE}(\alpha)$.

\subsection{Results}\label{sec:expt2_results}

\begin{figure}[]
\centering
\includegraphics[width=\columnwidth]{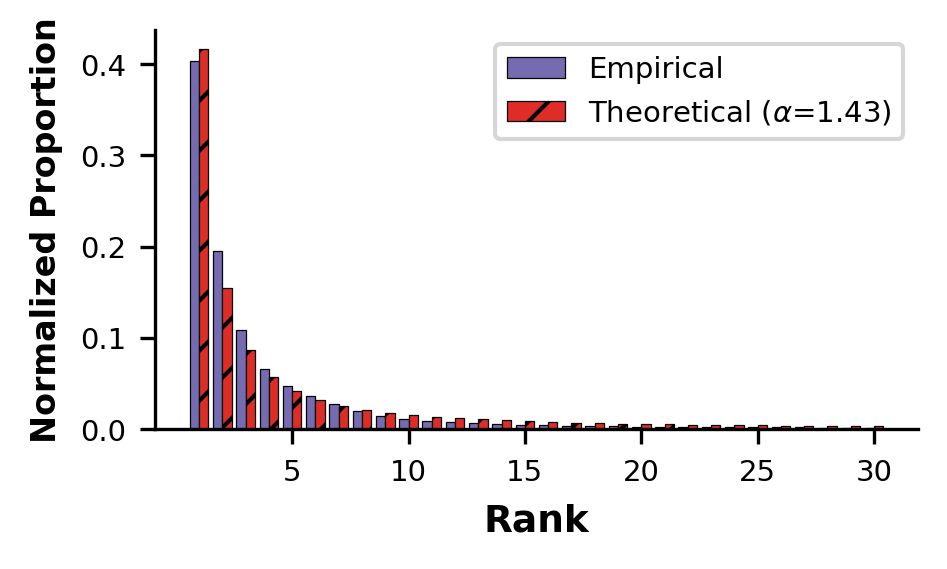}
\caption{The empirical distribution of subject-verb pairings in CHILDES (averaged across verbs in accordance with Equation~\ref{eq:empirical}), along with the frequencies predicted by a Zipfian distribution with parameter $\alpha=1.43$ ($\alpha$ was chosen by finding the best fit to the data). 
}
\label{fig:empirical_theoretical}
\end{figure}

\begin{figure}[t]
    \centering
    \includegraphics[width=0.9\columnwidth]{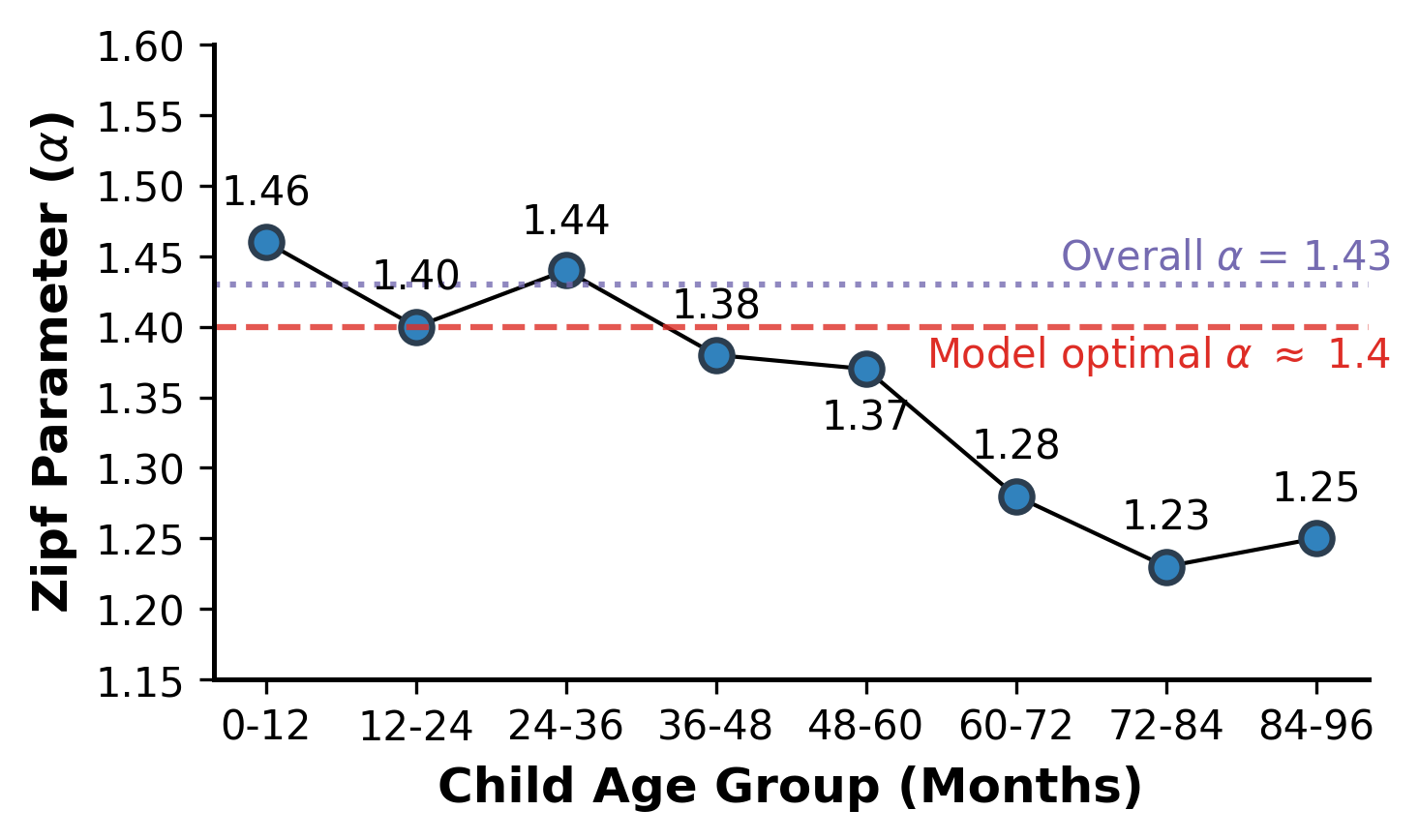}
    \caption{The fitted Zipf parameter $\alpha$ decreases with child age. The dashed red line indicates the optimal $\alpha$ found in neural network simulations; the dotted purple line indicates the overall corpus $\alpha$.}
    \label{fig:zipf_age}
\end{figure}

We found that the best-fitting value of $\alpha$ was $\alpha$=1.43. See Figure~\ref{fig:empirical_theoretical} for a comparison between $f_{\text{empirical}}$ and $f_{\text{theoretical}}$ with this $\alpha$ value. 
In addition to the dataset-wide fitting described above, we also broke down the analysis by the age of the child being spoken to in order to see whether the best-fitting $\alpha$ value varied by the age of the target child. As shown in Figure~\ref{fig:zipf_age}, the Zipfian parameter $\alpha$ generally decreases as the target child’s age increases. 
Sample utterances by age group are in Appendix~\ref{appendix:child_directed_speech}. 

Strikingly, both the $\alpha$ value calculated for all utterances ($\alpha$=1.43) and the range of $\alpha$ values found for each age group ($\alpha$=1.46 to 1.23)  are close to the value of $\alpha$ where our model generalized best, $\alpha \approx 1.4$. This finding suggests that naturalistic English input has a level
of subject-verb variability that facilitates the acquisition of agreement.


\section{Discussion}\label{sec:discussion}

We have used neural network language models to show that it is possible for statistical learners to robustly generalize English subject-verb agreement by using collocational bootstrapping. This bootstrapping strategy only succeeds under certain statistical conditions (when the $\alpha$ parameter in Zipf's law is about 1.4); we have further found preliminary evidence that child-directed speech has the right properties for this strategy to be viable.

\paragraph{Making inferences about child language acquisition:} Due to the many differences between our synthetic text and natural child-directed language, we do not intend to draw strong conclusions about the similarity between the model-optimal $\alpha$ value ($\approx$ 1.4) and the empirical $\alpha$ found in CHILDES (1.43). It is worth noting that the type of simulations we conducted, whether done with fully synthetic data or data closer to child-directed language, can only provide evidence about which learning strategies could be effective, not whether children actually use those strategies during acquisition. 

\paragraph{Toward greater naturalness:} Our synthetic data sets were highly simplified, differing from naturalistic language in important ways. First, our existing data sets likely over-represent the presence of prepositional phrases before the verb. Second, our training sets were fully ambiguous between \textsc{Agree-Subject} and \textsc{Agree-Recent} whereas naturalistic data contain some disambiguating examples---though naturalistic data can also contain agreement attraction errors \cite{bock1991broken} that point toward \textsc{Agree-Recent} rather than \textsc{Agree-Subject}. Third, naturalistic data involve a much larger vocabulary than what we used here. Fourth, naturalistic English sentences often use verbs (e.g., past-tense verbs) that are not explicitly inflected for number. 

Beyond these differences in the word sequences encountered by our models vs. human children, our models also differ from human learners in only having access to text, whereas children receive multi-modal input that might support types of bootstrapping not available to our models. Children have access to prosody, which might provide syntactic cues through prosodic bootstrapping, and can also draw on real-world context, which might provide meaning that can serve as a cue to syntax, as suggested under semantic bootstrapping. Future work could explore the effects of modifying the training set in ways that overcome these qualitative gaps.




\paragraph{The difficulty of agreement acquisition:} Our analysis of CHILDES found that its statistical properties make it well-suited for collocational bootstrapping. However, prior work has found that both children \cite{nozari2022revisiting} and neural networks trained on child-directed language \cite{huebner2021babyberta, Padovani_2025} make agreement attraction errors, meaning that they have not learned subject-verb agreement as robustly as might be expected from our analysis. 
A likely explanation for the discrepancy is that some of the other factors mentioned in the previous paragraph could counteract the favorable $\alpha$ value that we have observed in ways that add further difficulty to the acquisition task. A goal to ultimately work toward is investigating which types of input data and learning strategies can reproduce both the successes and failures of subject-verb agreement in humans.

\paragraph{Statistical co-occurrence or semantic relatedness?} Semantic bootstrapping leverages the meaning of words as a cue to syntax. 
Since semantically related words often occur near each other, 
there may be overlap between semantic and collocational cues. Indeed, past computational work has found a relationship between a word's meaning and its statistical distribution in a corpus  \cite{landauer1997solution,mikolov2013distributed}. Future work could tease apart the respective roles of semantics and statistics as cues to syntactic dependencies.

Semantic bootstrapping is typically framed as a mechanism for inferring syntactic categories such as parts of speech. Collocational bootstrapping is instead a strategy for acquiring word-word dependencies, under which learners can bootstrap from one type of word-word relatedness (co-occurrence) to another (syntactic dependencies). Since semantics and distribution overlap, this same broad strategy could instead use semantic relatedness rather than distributional co-occurrence as a cue to syntactic dependencies, providing a way to extend semantic bootstrapping to the learning of dependencies.

\paragraph{Extending collocational bootstrapping:} Another direction for future work is extending collocational bootstrapping by analyzing whether it is an effective strategy for learning other syntactic dependencies beyond the one studied here (subject-verb dependencies). 
A natural first step would be to investigate other agreement phenomena in English and other languages, such as noun-anaphor number agreement and adjective-noun gender agreement.


\section{Conclusion}

We have proposed collocational bootstrapping as a potential mechanism by which word co-occurrence statistics can support the learning of syntax. We have tested our hypothesis using neural language models, training and evaluating them on synthetic data with varying levels of variability in subject-verb pairings. We have found that there is an optimal level of variability, specifically a Zipfian distribution with $\alpha \approx 1.4$, that maximizes the model's ability to generalize. Too little variability prevents the model from generalizing to novel noun-verb pairs, and too much variability prevents it from abstracting syntactic rules. The $\alpha$ value at which there is a sweet spot for optimal generalization is consistent with the level of variability observed in child-directed speech ($\alpha$=1.43), suggesting that the statistical structure of natural language could guide learners in correctly acquiring syntax. 
These results provide one illustration of how statistical properties of linguistic data can facilitate the learning of abstract syntactic phenomena.



\section*{Limitations}

Our neural network experiments involve simplified, synthetic training data that differ from children's input in qualitative ways, and we have only analyzed the effect of one statistical cue on one linguistic phenomenon; see Section~\ref{sec:discussion} for discussion. 

\section*{Acknowledgments}
We extend our thanks to the anonymous reviewers and the feedback they provided on this paper, and to Jason Hobbs for his technical insight and support. We used Claude Code for assistance with coding, and we checked all AI-generated code. We also used Grammarly, and Claude Opus 4.6 and Sonnet 4.5, for feedback on style and grammar, but all ideas in the paper were ours. Any errors are our own.


\bibliography{custom}

\appendix

\section{Sample training sentences}
\label{appendix:training_sentences}

Below are examples of sentences used in training for a few levels of variability.

\vspace{1em}
\noindent \textit{Maximally Variable} ($\alpha$ = 0):
\begin{itemize}[nosep]
    \item the driver leads
    \item by the solver the challenger trades
    \item the dancers near the writers embezzle
    \item by the twirlers the painters near the singers navigate
\vspace{1em}
\end{itemize}

\noindent \textit{Moderately Variable} ($\alpha$ = 1.5):
\begin{itemize}[nosep]
    \item the hunters listen
    \item by the builder the twirler collapses
    \item by the swimmers the bridgers bridge
    \item near the miner the jumper near the painter jumps
\vspace{1em}
\end{itemize}

\noindent \textit{No Variability} ($\alpha \to \infty$):
\begin{itemize}[nosep]
    \item the twirler twirls
    \item by the charmers the miners mine
    \item the builders near the protectors build
    \item near the swimmers the lassoers by the bakers lasso
\vspace{1em}
\end{itemize}

\section{Speaker role utterance counts}
\label{appendix:speaker_role_counts}
\begin{figure}[t]
    \centering
    \includegraphics[width=\columnwidth]{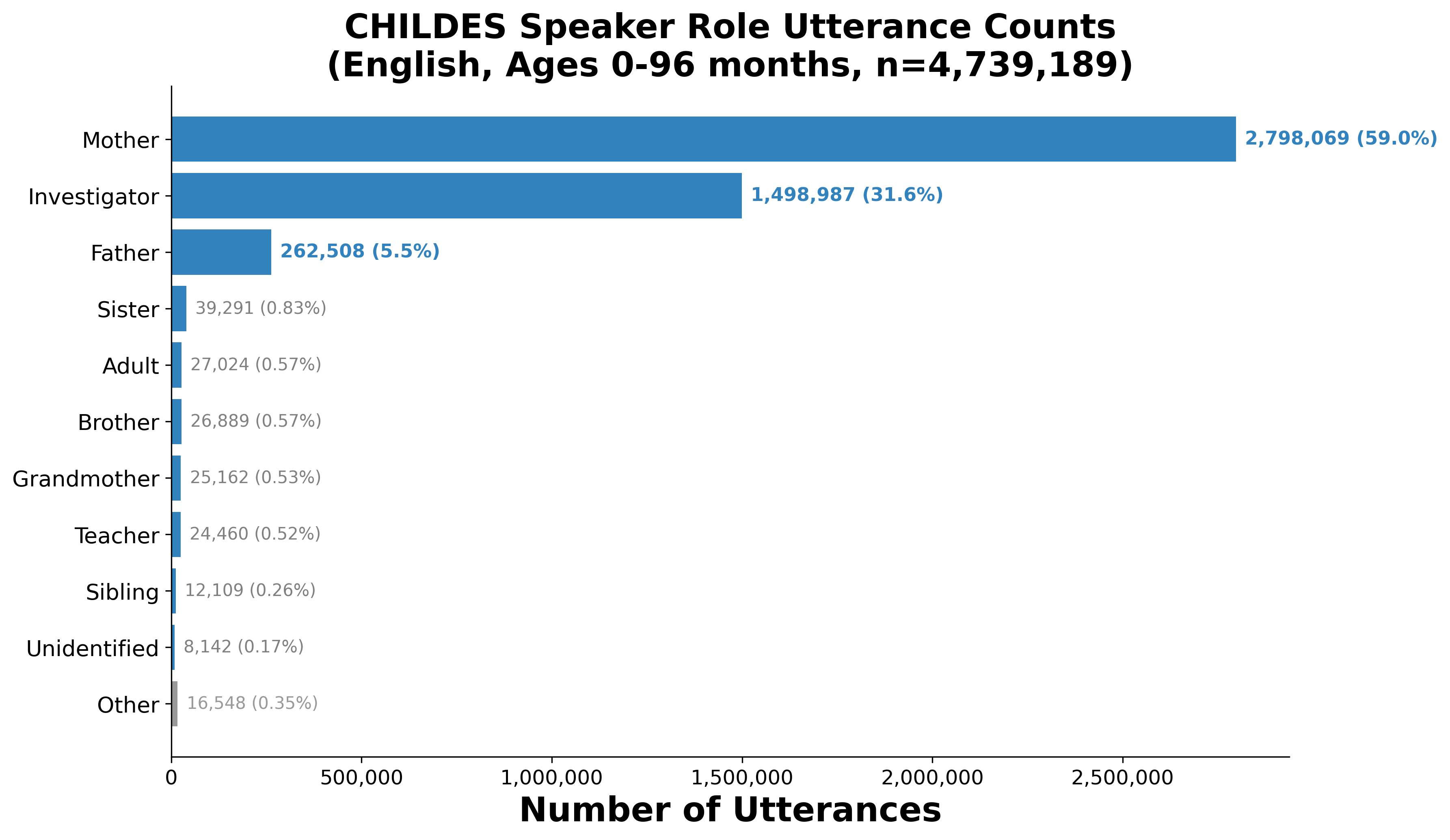}
    \caption{Distribution of utterances by speaker role in the English subset of CHILDES (ages 0-96 months).}
    \label{fig:speaker_roles_96mos}
\end{figure}

See Figure~\ref{fig:speaker_roles_96mos} for utterance counts by each type of speaker in the CHILDES data that we analyzed.

\section{Examples of child-directed language at varying child ages}
\label{appendix:child_directed_speech}

Below are examples of child-directed language spoken to children of varying ages.

\bigskip
\noindent Age 0-12 months ($\alpha = 1.46$)
\begin{enumerate}[nosep]
    \item ``you put the block on''
    \item ``what else do we see in here''
    \item ``oh be be very gentle with baby right''
\end{enumerate}

\vspace{1em}
\noindent Age 12-24 months ($\alpha = 1.40$)
\begin{enumerate}[nosep]
    \item ``what's that''
    \item ``yeah that's where we were''
    \item ``you don't like MacDonald's and I don't like MacDonald's''
\end{enumerate}

\vspace{1em}
\noindent Age 24-36 months ($\alpha = 1.44$)
\begin{enumerate}[nosep]
    \item ``how do you know this is a duck''
    \item ``this is velcro''
    \item ``let's sit here on mama's mama's knee''
\end{enumerate}

\vspace{1em}
\noindent Age 36-48 months ($\alpha = 1.38$)
\begin{enumerate}[nosep]
    \item ``you get milk from it''
    \item ``look at these''
    \item ``want mommy to read''
\end{enumerate}

\vspace{1em}
\noindent Age 48-60 months ($\alpha = 1.37$)
\begin{enumerate}[nosep]
    \item ``i think we found the wheels or your mom did''
    \item ``just like we see up there remember''
    \item ``there's something wrong with her teeth aren't there''
\end{enumerate}

\vspace{1em}
\noindent Age 60-72 months ($\alpha = 1.28$)
\begin{enumerate}[nosep]
    \item ``well I know but you know what I think this chair is''
    \item ``so you want listen come here I'm going to tell you''
    \item ``I don't think I would like those''
\end{enumerate}

\vspace{1em}
\noindent Age 70-84 months ($\alpha = 1.23$)
\begin{enumerate}[nosep]
    \item ``I never heard of that one before''
    \item ``dad's gonna dads can do it a lot''
    \item ``there's how many bears on one wheel'
\end{enumerate}

\vspace{1em}
\noindent Age 84-96 months ($\alpha = 1.25$)
\begin{enumerate}[nosep]
    \item ``I got you a pencil''
    \item ``alright well if you don't put it on then the letter's no good''
    \item ``uh what about a movie though''
\end{enumerate}

\section{Training loss}
\label{appendix:training_loss}
\begin{figure*}[t]
    \centering
    \includegraphics[width=\textwidth]{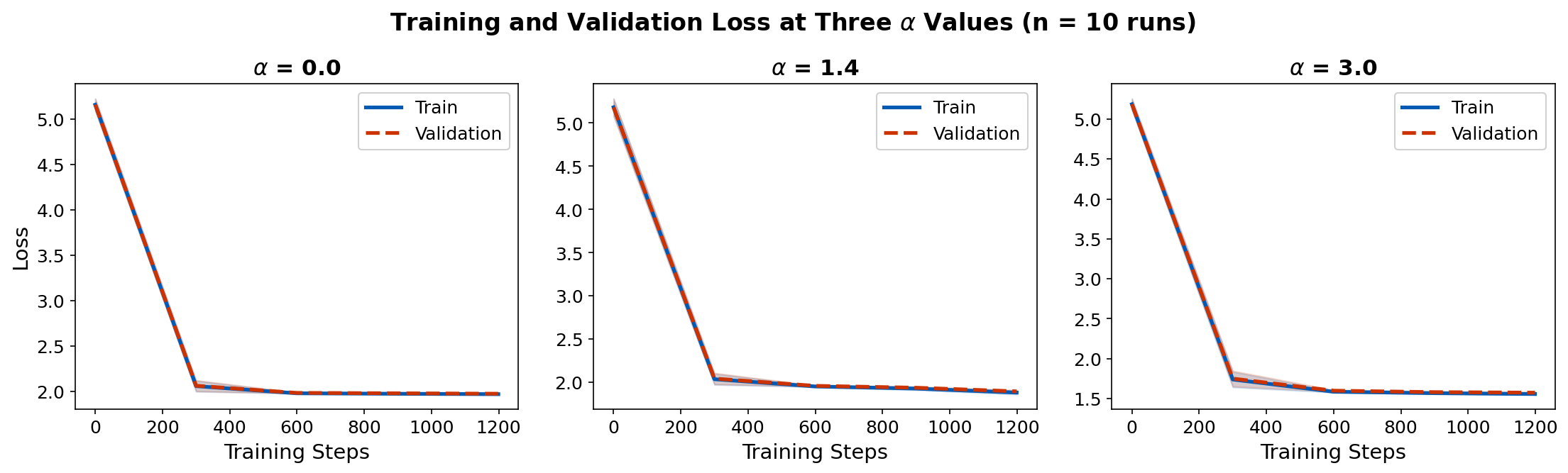}
    \caption{Training and validation loss at three $\alpha$ values. Loss curves track closely across all conditions, indicating no overfitting.}
    \label{fig:loss}
\end{figure*}

See Figure~\ref{fig:loss} for the loss trajectories of the models we trained.

\section{Data Cleaning}
\label{appendix:data_cleaning}
We downloaded 5,147,586 utterances from participants categorized as English-language speakers, restricted to 25 target speaker roles: Adult, Caretaker, Father, Friend, Grandfather, Grandmother, Investigator, Mother, Narrator, Playmate, Relative, Sibling, Sister, Brother, Teacher, Unidentified, Visitor, Teenager, Participant, Girl, Male, Student, Environment, Doctor, Target Adult. Next, we removed rows with null text content, converted utterances to text strings, removed rows lacking a target child age, and removed rows where the target child age was greater than 96 months.
After cleaning, 4,739,189 utterances remained. Of these, 59.0\% were spoken by mothers and 31.6\% by investigators, together comprising nearly 90\% of all utterances as shown in Figure~\ref{fig:speaker_roles_96mos} in Appendix~\ref{appendix:speaker_role_counts}. This proportion reflects the high concentration of speech from caregivers in the corpus \cite{kempe_2024}. During the subject-verb extraction step, subject and verb lemmas were converted to lower case, and the resulting verb counts were restricted to ASCII English forms before selecting the top 100 verbs for analysis.

\section{Analysis of subject-verb pairings by age}
See Table~\ref{tab:age_groups} for statistics of subject-verb pairings in child-directed language broken down by the age of the child being spoken to.

\label{appendix:pairings_by_age}
\begin{table*}[]
\centering
\begin{tabular}{lrrrrrr}
\hline
\textbf{Age Group} & \textbf{Utterances} & \textbf{S-V Pairs} & \textbf{Unique Verbs} & \textbf{Unique Subjects} & \textbf{$\alpha$} & \textbf{MSE} \\
\hline
0--12mo   & 182,023   & 113,607   & 1,180  & 1,516  & 1.46 & 9.78e-07 \\
12--24mo  & 671,559   & 350,859   & 3,580  & 6,135  & 1.40 & 2.56e-07 \\
24--36mo  & 1,900,684 & 1,163,974 & 6,775  & 13,577 & 1.44 & 2.15e-07 \\
36--48mo  & 923,858   & 553,675   & 4,940  & 9,817  & 1.38 & 3.90e-07 \\
48--60mo  & 621,805   & 399,355   & 4,262  & 8,022  & 1.37 & 5.26e-07 \\
60--72mo  & 237,449   & 121,409   & 2,661  & 4,805  & 1.28 & 2.33e-07 \\
72--84mo  & 111,117   & 52,385    & 1,778  & 3,092  & 1.23 & 7.79e-07 \\
84--96mo  & 90,694    & 46,807    & 1,467  & 2,491  & 1.25 & 4.87e-07 \\
\hline
\end{tabular}
\caption{Age-stratified analysis of subject-verb pairings in CHILDES (0--96 months). The Zipf parameter $\alpha$ decreases from 1.46 in the youngest age group to 1.25 in the oldest.}
\label{tab:age_groups}
\end{table*}

\end{document}